\documentclass[10pt,twocolumn,letterpaper]{article}

\usepackage{iccv}
\usepackage{times}
\usepackage{epsfig}
\usepackage{graphicx}
\usepackage{amsmath}
\usepackage{amssymb}
\usepackage{mathtools}
\usepackage{multirow}
\usepackage{enumitem}
\usepackage{caption}
\usepackage{subcaption}
\usepackage{array}
\usepackage{colortbl}
\usepackage{booktabs}
\usepackage{bbm}
\usepackage{multicol}
\usepackage{makecell}
\usepackage{xcolor}
\usepackage{xspace}
\usepackage{float}
\usepackage{color}
\usepackage{pifont}
\setitemize[1]{itemsep=0pt,partopsep=0pt,parsep=\parskip,topsep=5pt}

\usepackage[pagebackref=true,breaklinks=true,letterpaper=true,colorlinks,bookmarks=false]{hyperref}

\iccvfinalcopy

\def\iccvPaperID{****}

\ificcvfinal\fi

\newlength\savedwidth
\newcommand\whline{\noalign{\global\savedwidth\arrayrulewidth\global\arrayrulewidth 0.8pt}\hline\noalign{\global\arrayrulewidth\savedwidth}}

\newcommand{\tabincell}[2]{\begin{tabular}{@{}#1@{}}#2\end{tabular}}
\definecolor{mygray}{gray}{.92}

\def\ie{\emph{i.e.}}
\def\eg{\emph{e.g.}}

\begin{document}
	
	\title{PVT v2: Improved Baselines with Pyramid Vision Transformer}
	
	\author{
		Wenhai Wang$^{1,2}$, 
		Enze Xie$^{3}$,
		Xiang Li$^{4}$,
		Deng-Ping Fan$^{5}$,\\
		Kaitao Song$^{4}$,
		Ding Liang$^{6}$,
		Tong Lu$^{2}$,
		Ping Luo$^{3}$,
		Ling Shao$^{5}$\\
		$^1$Shanghai AI Laboratory~~~
		$^2$Nanjing University~~~
		$^3$The University of Hong Kong~~~\\
		$^4$Nanjing University of Science and Technology~~~
		$^5$IIAI~~~
		$^6$SenseTime Research\\
		$\tt wangwenhai@pjlab.org.cn$,
	}
	
	\maketitle
	\ificcvfinal\thispagestyle{empty}\fi

	\begin{abstract}
		Transformer recently has presented encouraging progress in computer vision. 
		In this work, we present new baselines by improving the original Pyramid Vision Transformer (PVT v1) by adding three designs, including 
		(1) linear complexity attention layer, (2) overlapping patch embedding, and (3) convolutional feed-forward network.
		With these modifications, PVT v2 reduces the computational complexity of PVT v1 to linear and achieves significant improvements on fundamental vision tasks such as classification, detection, and segmentation.
		Notably, the proposed PVT v2 achieves comparable or better performances than recent works such as Swin Transformer.
		We hope this work will facilitate state-of-the-art  Transformer researches in computer vision. Code is available at {\url{https://github.com/whai362/PVT}}.
	\end{abstract}

	\section{Introduction}\label{sec:introduction}
	
	Recent studies on vision Transformer are converging on the backbone network~\cite{dosovitskiy2020image,touvron2020training,pvt,cvt,swin,coat,levit,twins} designed for downstream vision tasks, such as image classification, object detection, instance and semantic segmentation.
	To date, there have been some promising results. 
	For example, Vision Transformer (ViT)~\cite{dosovitskiy2020image} first proves that a pure Transformer can archive state-of-the-art performance in image classification.
	Pyramid Vision Transformer (PVT v1)~\cite{pvt} shows that a pure Transformer backbone can also surpass CNN counterparts in dense prediction tasks such as detection and segmentation tasks~\cite{lin2014microsoft,zhou2017scene}.
	After that, Swin Transformer~\cite{swin}, CoaT~\cite{coat}, LeViT~\cite{levit}, and Twins~\cite{twins} further improve the classification, detection, and segmentation performance with Transformer backbones.
	
	This work aims to establish stronger and more feasible baselines built on the PVT v1 framework.
	We report that three design improvements, namely (1) linear complexity attention layer, (2) overlapping patch embedding, and (3) convolutional feed-forward network
	are orthogonal to the PVT v1 framework,
	and when used with PVT v1, they can bring better image classification, object detection, instance and semantic segmentation performance.
	The improved framework is termed as PVT v2.
	Specifically, PVT v2-B5\footnote{
		PVT v2 has 6 different size variants, from B0 to B5 according to the parameter number.}
	yields 83.8\% top-1 error on ImageNet, which is better than Swin-B~\cite{swin} and Twins-SVT-L~\cite{twins}, while our model has fewer parameters and GFLOPs.
	Moreover, GFL~\cite{li2020generalized} with PVT-B2 archives 50.2 AP on COCO \texttt{val2017}, 2.6 AP higher than the one with Swin-T~\cite{swin}, 5.7 AP higher than the one with ResNet50~\cite{he2015delving}.
	We hope these improved baselines will provide a reference
	for future research in vision Transformer.
	
	\section{Related Work} 
	
	We mainly discuss transformer backbones related to this work.
	ViT~\cite{dosovitskiy2020image} treats each image as a sequence of tokens (patches) with a fixed length, and then feeds them to multiple Transformer layers to perform classification.
	It is the first work to prove that a pure Transformer can also archive state-of-the-art performance in image classification when training data is sufficient (\eg, ImageNet-22k~\cite{deng2009imagenet}, JFT-300M).
	DeiT~\cite{touvron2020training} further explores a data-efficient training strategy and a distillation approach for ViT.

	To improve image classification performance, recent methods make tailored changes to ViT.
	T2T ViT~\cite{t2tvit} concatenates tokens within an overlapping sliding window into one token progressively.
	TNT~\cite{tnt} utilizes inner and outer Transformer blocks to generate pixel and patch embeddings respectively.
	CPVT~\cite{cpvt} replaces the fixed size position embedding in ViT with conditional position encodings, making it easier to process images of arbitrary resolution.
	CrossViT~\cite{crossvit} processes image patches of different sizes via a dual-branch Transformer.
	LocalViT~\cite{localvit} incorporates depth-wise convolution into vision Transformers to improve the local continuity of features.
	
	To adapt to dense prediction tasks such as object detection, instance and semantic segmentation, there are also some methods~\cite{pvt,swin,cvt,coat,levit,twins} to introduce the pyramid structure in CNNs to the design of Transformer backbones.
	PVT v1 is the first pyramid structure Transformer, which presents a hierarchical Transformer with four stages, showing that a pure Transformer backbone can be as versatile as CNN counterparts and performs better in detection and segmentation tasks.
	After that, some improvements~\cite{swin,cvt,coat,levit,twins} are made to enhance the local continuity of features and to remove fixed size position embedding.
	For example, Swin Transformer~\cite{swin} replaces fixed size position embedding with relative position biases, and restricts self-attention within shifted windows.
	CvT~\cite{cvt}, CoaT~\cite{coat}, and LeViT~\cite{levit} introduce convolution-like operations into vision Transformers.
	Twins~\cite{twins} combines local attention and global attention mechanisms to obtain stronger feature representation.

	\section{Methodology}
	
	\subsection{Limitations in PVT v1}
	There are three main limitations in PVT v1~\cite{pvt} as follows:
	(1) Similar to ViT~\cite{dosovitskiy2020image}, when processing high-resolution input (\eg,  shorter side being 800 pixels), the computational complexity of PVT v1 is relatively large.
	(2) PVT v1~\cite{pvt} treats an image as a sequence of non-overlapping patches, which loses the local continuity of the image to a certain extent;
	(3) The position encoding in PVT v1 is fixed-size, which is inflexible for process images of arbitrary size.
	These problems limit the performance of PVT v1 on vision tasks.
	
	To address these issues, we propose PVT v2, which improves PVT v1 through three designs, which are listed in Sec \ref{sec:l-sra}, \ref{sec:o-pe}, and \ref{sec:c-ffn}.
	
	\begin{figure}[t]
		\centering
		\setlength{\fboxrule}{0pt}
		\fbox{\includegraphics[width=0.45\textwidth]{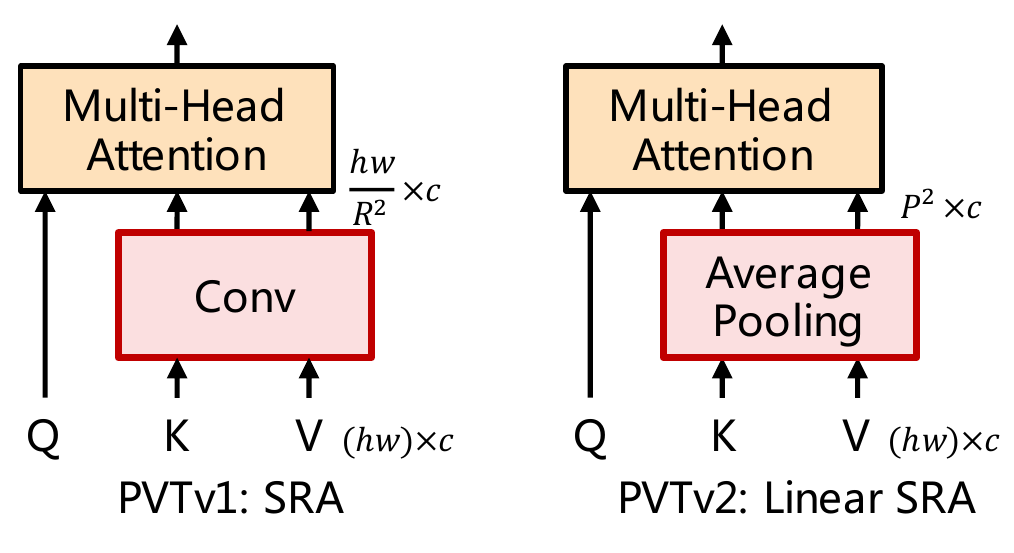}}
		\caption{\textbf{Comparison of SRA in PVT v1 and linear SRA in PVT v2.}}
		\label{fig:li_att}
	\end{figure}
	
	\subsection{Linear Spatial Reduction Attention}
	\label{sec:l-sra}
	First, to reduce the high computational cost caused by attention operations, we propose linear spatial reduction attention (SRA) layer as illustrated in Fig. \ref{fig:li_att}.
	Different from SRA~\cite{pvt} which uses convolutions for spatial reduction, linear SRA uses average pooling to reduce the spatial dimension (\ie, $h\times w$) to a fixed size (\ie, $P\times P$) before the attention operation.
	So linear SRA enjoys linear computational and memory costs like a convolutional layer.
	Specifically, given an input of size $h\times w \times c$, the complexity of SRA and linear SRA are:
	\begin{equation}
	\Omega({\rm SRA}) = \frac{2h^2w^2c}{R^2} + hwc^2R^2,
	\label{eqn:sra}
	\end{equation}
	\begin{equation}
	\Omega({\rm Linear\ SRA}) = 2hwP^2c,
	\label{eqn:lisra}
	\end{equation}
	where $R$ is the spatial reduction ratio of SRA~\cite{pvt}. $P$ is the pooling size of linear SRA, which is set to 7.
	
	\begin{figure}[t]
		\centering
		\setlength{\fboxrule}{0pt}
		\fbox{\includegraphics[width=0.48\textwidth]{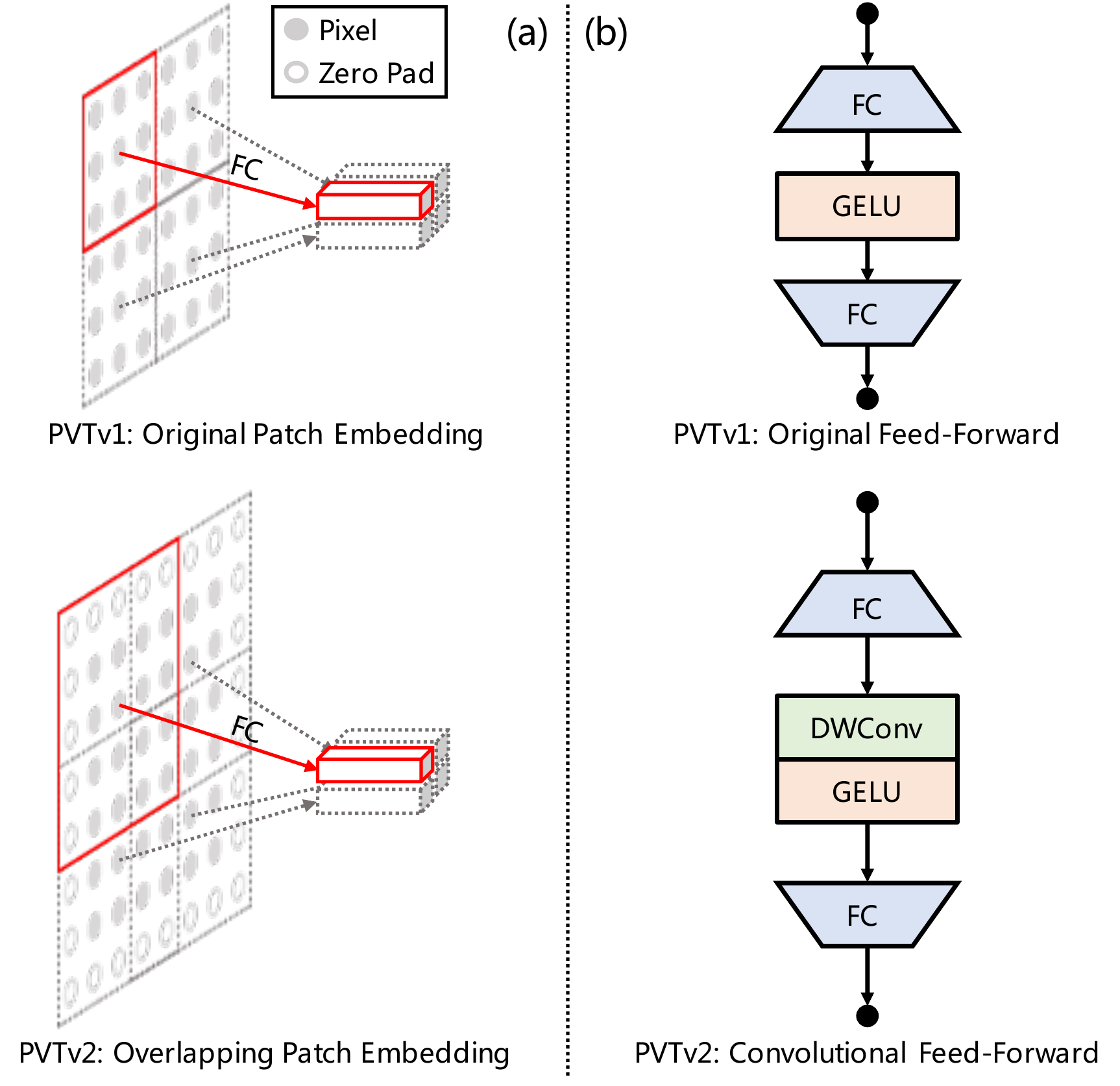}}
		\caption{\textbf{Two improvements in PVT v2.} (1) Overlapping Patch Embedding. (2) Convolutional Feed Forward Network.}
		\label{fig:diff}
	\end{figure}
	
	\subsection{Overlapping Patch Embedding} 
	\label{sec:o-pe}
	Second, to model the local continuity information, we utilize overlapping patch embedding to tokenize images. As shown in Fig. \ref{fig:diff}(a), we enlarge the patch window, making adjacent windows overlap by half of the area, and pad the feature map with zeros to keep the resolution.
	In this work, we use convolution with zero paddings to implement overlapping patch embedding.
	Specifically, given an input of size $h\times w\times c$, we feed it to a convolution with the stride of $S$, the kernel size of $2S - 1$, the padding size of $S - 1$, and the kernel number of $c'$.
	The output size is $\frac{h}{S}\times \frac{w}{S}\times C'$.
	
	\begin{table*}[t]
		\scriptsize
		\centering
		\begin{tabular}{c|c|c|c|c|c|c|c|c|c}
 & \multirow{2}{*}{Output Size} & \multirow{2}{*}{Layer Name} & \multicolumn{7}{c}{Pyramid Vision Transformer v2} \\
\cline{4-10}
 & & & B0 & B1 & B2 & B2-Li & B3 & B4 & B5\\
\hline
\multirow{3}{*}[-3.5ex]{Stage 1} & \multirow{3}{*}[-3.5ex]{\scalebox{1.3}{$\frac{H}{4}\times \frac{W}{4}$}} & \multirow{2}{*}{\tabincell{c}{Overlapping\\Patch Embedding}} &\multicolumn{7}{c}{$S_1=4$} \\
\cline{4-10}
& & & $C_1=32$ & \multicolumn{6}{c}{$C_1=64$}\\
\cline{3-10}
& & \tabincell{c}{Transformer\\Encoder} & 
$\begin{matrix}
R_1=8 \\
N_1=1 \\
E_1=8 \\
L_1=2 \\
\end{matrix}$ &
$\begin{matrix}
R_1=8 \\
N_1=1 \\
E_1=8 \\
L_1=2 \\
\end{matrix}$ &
$\begin{matrix}
R_1=8 \\
N_1=1 \\
E_1=8 \\
L_1=3 \\
\end{matrix}$ &
$\begin{matrix}
P_1=7 \\
N_1=1 \\
E_1=8 \\
L_1=3 \\
\end{matrix}$ &
$\begin{matrix}
R_1=8 \\
N_1=1 \\
E_1=8 \\
L_1=3 \\
\end{matrix}$ &
$\begin{matrix}
R_1=8 \\
N_1=1 \\
E_1=8 \\
L_1=3 \\
\end{matrix}$ &
$\begin{matrix}
R_1=8 \\
N_1=1 \\
E_1=4 \\
L_1=3 \\
\end{matrix}$\\
\hline
\multirow{3}{*}[-3.5ex]{Stage 2} & \multirow{3}{*}[-3.5ex]{\scalebox{1.3}{$\frac{H}{8}\times \frac{W}{8}$}} & \multirow{2}{*}{\tabincell{c}{Overlapping\\Patch Embedding}} & \multicolumn{7}{c}{$S_2=2$} \\
\cline{4-10}
& & & $C_2=64$ & \multicolumn{6}{c}{$C_2=128$}\\
\cline{3-10}
& & \tabincell{c}{Transformer\\Encoder} &
$\begin{matrix}
R_2=4 \\
N_2=2 \\
E_2=8 \\
L_2=2 \\
\end{matrix}$ &
$\begin{matrix}
R_2=4 \\
N_2=2 \\
E_2=8 \\
L_2=2 \\
\end{matrix}$ &
$\begin{matrix}
R_2=4 \\
N_2=2 \\
E_2=8 \\
L_2=3 \\
\end{matrix}$ &
$\begin{matrix}
P_2=7 \\
N_2=2 \\
E_2=8 \\
L_2=3 \\
\end{matrix}$ &
$\begin{matrix}
R_2=4 \\
N_2=2 \\
E_2=8 \\
L_2=3 \\
\end{matrix}$ &
$\begin{matrix}
R_2=4 \\
N_2=2 \\
E_2=8 \\
L_2=8 \\
\end{matrix}$ &
$\begin{matrix}
R_2=4 \\
N_2=2 \\
E_2=4 \\
L_2=6 \\
\end{matrix}$\\
\hline
\multirow{3}{*}[-3.5ex]{Stage 3} & \multirow{3}{*}[-3.5ex]{\scalebox{1.3}{$\frac{H}{16}\times \frac{W}{16}$}} & \multirow{2}{*}{\tabincell{c}{Overlapping\\Patch Embedding}} & \multicolumn{7}{c}{$S_3=2$} \\
\cline{4-10}
& & & $C_3=160$ & \multicolumn{6}{c}{$C_3=320$}\\
\cline{3-10}
& & \tabincell{c}{Transformer\\Encoder} &
$\begin{matrix}
R_3=2 \\
N_3=5 \\
E_3=4 \\
L_3=2 \\
\end{matrix}$ &
$\begin{matrix}
R_3=2 \\
N_3=5 \\
E_3=4 \\
L_3=2 \\
\end{matrix}$ &
$\begin{matrix}
R_3=2 \\
N_3=5 \\
E_3=4 \\
L_3=6 \\
\end{matrix}$ &
$\begin{matrix}
P_3=7 \\
N_3=5 \\
E_3=4 \\
L_3=6 \\
\end{matrix}$ &
$\begin{matrix}
R_3=2 \\
N_3=5 \\
E_3=4 \\
L_3=18 \\
\end{matrix}$ &
$\begin{matrix}
R_3=2 \\
N_3=5 \\
E_3=4 \\
L_3=27 \\
\end{matrix}$ &
$\begin{matrix}
R_3=2 \\
N_3=5 \\
E_3=4 \\
L_3=40 \\
\end{matrix}$\\
\hline
\multirow{3}{*}[-3.5ex]{Stage 4} &  \multirow{3}{*}[-3.5ex]{\scalebox{1.3}{$\frac{H}{32}\times \frac{W}{32}$}} & \multirow{2}{*}{\tabincell{c}{Overlapping\\Patch Embedding}} & \multicolumn{7}{c}{$S_4=2$} \\
\cline{4-10}
& & & $C_4=256$ & \multicolumn{6}{c}{$C_4=512$}\\
\cline{3-10}
& & \tabincell{c}{Transformer\\Encoder} & 
$\begin{matrix}
R_4=1 \\
N_4=8 \\
E_4=4 \\
L_4=2 \\
\end{matrix}$ &
$\begin{matrix}
R_4=1 \\
N_4=8 \\
E_4=4 \\
L_4=2 \\
\end{matrix}$ &
$\begin{matrix}
R_4=1 \\
N_4=8 \\
E_4=4 \\
L_4=3 \\
\end{matrix}$ & 
$\begin{matrix}
P_4=7 \\
N_4=8 \\
E_4=4 \\
L_4=3 \\
\end{matrix}$ & 
$\begin{matrix}
R_4=1 \\
N_4=8 \\
E_4=4 \\
L_4=3 \\
\end{matrix}$ & 
$\begin{matrix}
R_4=1 \\
N_4=8 \\
E_4=4 \\
L_4=3 \\
\end{matrix}$ &
$\begin{matrix}
R_4=1 \\
N_4=8 \\
E_4=4 \\
L_4=3 \\
\end{matrix}$\\
\end{tabular}

		\caption{\textbf{Detailed settings of PVT v2 series.} ``-Li'' denotes PVT v2 with linear SRA.} 
		\label{tab:arch}
	\end{table*}
	
	\subsection{Convolutional Feed-Forward} 
	\label{sec:c-ffn}
	Third, inspired by \cite{zeropadding,cpvt,localvit}, we remove the fixed-size position encoding~\cite{dosovitskiy2020image}, and introduce zero padding position encoding into PVT.
	As shown in Fig. \ref{fig:diff}(b), we add a $3\times 3$ depth-wise convolution~\cite{howard2017mobilenets} with the padding size of 1 between the first fully-connected (FC) layer and GELU~\cite{gelu} in feed-forward networks.

	\subsection{Details of PVT v2 Series}
	We scale up PVT v2 from B0 to B5 By changing the hyper-parameters.
	which are listed as follows:
	\begin{itemize}
		\item $S_i$: the stride of the overlapping patch embedding in Stage $i$;
		\item $C_i$: the channel number of the output of Stage $i$;
		\item $L_i$: the number of encoder layers in Stage $i$;
		\item $R_i$: the reduction ratio of the SRA in Stage $i$;
		\item $P_i$: the adaptive average pooling size of the linear SRA in Stage $i$;
		\item $N_i$: the head number of the Efficient Self-Attention in Stage $i$;
		\item $E_i$: the expansion ratio of the feed-forward layer~\cite{vaswani2017attention} in Stage $i$;
	\end{itemize}
	
	Tab.~\ref{tab:arch} shows the detailed information of PVT v2 series. Our design follows the principles of ResNet~\cite{he2016deep}. (1) the channel dimension increase while the spatial resolution shrink with the layer goes deeper. (2) Stage 3 is assigned to most of the computation cost.
	
	\subsection{Advantages of PVT v2} 
	Combining these improvements, PVT v2 can (1) obtain more local continuity of images and feature maps; (2) process variable-resolution input more flexibly; (3) enjoy the same linear complexity as CNN.
	
	\begin{table}[t]
		\centering
		\setlength{\tabcolsep}{1.1mm}
		\footnotesize
		\scalebox{1.0}{
			\begin{tabular}{l|c|c|c}
    \renewcommand{\arraystretch}{0.1}
	Method & \#Param (M) & GFLOPs & Top-1 Acc (\%)  \\
	\whline
	PVTv2-B0 (ours)  & \textbf{3.4} & \textbf{0.6} & \textbf{70.5} \\
	\hline
	ResNet18~\cite{he2016deep} & 11.7 & 1.8 & 69.8  \\
	DeiT-Tiny/16~\cite{touvron2020training} & \textbf{5.7} & \textbf{1.3} & 72.2 \\
	PVTv1-Tiny~\cite{pvt} & 13.2 & 1.9 &75.1 \\
	PVTv2-B1 (ours)  & 13.1 & 2.1 & \textbf{78.7} \\
	\hline
	ResNet50~\cite{he2016deep}   &25.6 &4.1 &76.1 \\
	ResNeXt50-32x4d~\cite{xie2017aggregated} &25.0 &4.3 &77.6  \\
	RegNetY-4G~\cite{regnet} & 21.0 & 4.0 & 80.0 \\
	DeiT-Small/16~\cite{touvron2020training}  & 22.1 & 4.6 & 79.9 \\
	T2T-ViT$_t$-14~\cite{t2tvit} & 22.0 & 6.1 & 80.7 \\
	PVTv1-Small~\cite{pvt}  & 24.5 & 3.8 & 79.8 \\
	TNT-S~\cite{tnt} & 23.8 & 5.2 & 81.3 \\
	Swin-T~\cite{swin} & 29.0 & 4.5 & 81.3 \\
	CvT-13~\cite{cvt} & \textbf{20.0} & 4.5 & 81.6 \\
	CoaT-Lite Small~\cite{coat} & \textbf{20.0} & 4.0 & 81.9 \\
	Twins-SVT-S~\cite{twins} & 24.0 & \textbf{2.8} & 81.7 \\
	PVTv2-B2-Li (ours) & 22.6 &3.9 & \textbf{82.1} \\
	PVTv2-B2 (ours) & 25.4 & 4.0 & 82.0 \\
	\hline
	ResNet101~\cite{he2016deep}  &44.7 & 7.9 &77.4\\
	ResNeXt101-32x4d~\cite{xie2017aggregated} & 44.2 & 8.0 &78.8 \\
	RegNetY-8G~\cite{regnet} & 39.0 & 8.0 & 81.7 \\
	T2T-ViT$_t$-19~\cite{t2tvit} & 39.0 & 9.8 & 81.4 \\
	PVTv1-Medium~\cite{pvt} & 44.2 & 6.7 & 81.2\\
	CvT-21~\cite{cvt} & \textbf{32.0} & 7.1 & 82.5 \\
	PVTv2-B3 (ours) & 45.2 & \textbf{6.9} & \textbf{83.2} \\
	\hline
	ResNet152~\cite{he2016deep} & 60.2 & 11.6 & 78.3 \\
	T2T-ViT$_t$-24~~\cite{t2tvit} & 64.0 & 15.0 & 82.2 \\
	PVTv1-Large~~\cite{pvt} & 61.4 & 9.8 & 81.7 \\
	TNT-B~~\cite{tnt} & 66.0 & 14.1 & 82.8 \\
	Swin-S~~\cite{swin} & \textbf{50.0} & 8.7 & 83.0 \\
	Twins-SVT-B~\cite{twins} & 56.0 & \textbf{8.3} & 83.2 \\
	PVTv2-B4 (ours) & 62.6 & 10.1 & \textbf{83.6} \\
	\hline
	ResNeXt101-64x4d~\cite{xie2017aggregated} & 83.5 & 15.6 & 79.6\\
	RegNetY-16G~\cite{regnet} & 84.0 & 16.0 & 82.9\\ 
	ViT-Base/16~\cite{dosovitskiy2020image} & 86.6 & 17.6 & 81.8 \\
	DeiT-Base/16~\cite{touvron2020training} & 86.6 & 17.6 & 81.8 \\
	Swin-B~\cite{swin} & 88.0 & 15.4 & 83.3 \\
	Twins-SVT-L~\cite{twins} & 99.2 & 14.8 & 83.7 \\
	PVTv2-B5 (ours) & \textbf{82.0} & \textbf{11.8} & \textbf{83.8}\\
\end{tabular}}
		\caption{\textbf{Image classification performance on the ImageNet validation set}.
			``\#Param'' refers to the number of parameters. 
			``GFLOPs'' is calculated under the input scale of $224\times 224$. ``*'' indicates the performance of the method trained under the strategy of its original paper.
			``-Li'' denotes PVT v2 with linear SRA.}
		\label{tab:cls}
	\end{table}
	
	\section{Experiment}
	\subsection{Image Classification}
	\noindent\textbf{Settings.}
	Image classification experiments are performed on the ImageNet-1K dataset \cite{russakovsky2015imagenet}, which comprises 1.28 million training images and 50K validation images from 1,000 categories. 
	All models are trained on the training set for fair comparison and report the top-1 error on the validation set.
	We follow DeiT~\cite{touvron2020training} and apply random cropping, random horizontal flipping \cite{szegedy2015going}, label-smoothing regularization \cite{szegedy2016rethinking}, mixup~\cite{zhang2017mixup}, and random erasing~\cite{zhong2020random} as data augmentations.
	During training, we employ AdamW~\cite{loshchilov2017decoupled} with a momentum of 0.9, a mini-batch size of 128, and a weight decay of $5\times 10^{-2}$ to optimize models. The initial learning rate is set to $1\times 10^{-3}$ and decreases following the cosine schedule~\cite{loshchilov2016sgdr}.  All models are trained for 300 epochs from scratch on 8 V100 GPUs.
	We apply a center crop on the validation set to benchmark, where a 224$\times$ 224 patch is cropped to evaluate the classification accuracy.

	\noindent\textbf{Results.}
	In Tab.~\ref{tab:cls}, we see that PVT v2 is the state-of-the-art method on ImageNet-1K classification.
	Compared to PVT, PVT v2 has similar flops and parameters, but the image classification accuracy is greatly improved. For example, PVT v2-B1 is 3.6\% higher than PVT v1-Tiny, and PVT v2-B4 is 1.9\% higher than PVT-Large.
	
	Compared to other recent counterparts, PVT v2 series also has large advantages in terms of accuracy and model size. For example, PVT v2-B5 achieves 83.8\% ImageNet top-1 accuracy, which is 0.5\% higher than Swin Transformer~\cite{swin} and Twins~\cite{twins}, while our parameters and FLOPS are fewer.

	\subsection{Object Detection}
	\noindent\textbf{Settings.}
	Object detection experiments are conducted on the challenging COCO benchmark~\cite{lin2014microsoft}.
	All models are trained on COCO \texttt{train2017} (118k images) and evaluated on \texttt{val2017} (5k images).
	We verify the effectiveness of PVT v2 backbones on top of mainstream detectors, including RetinaNet~\cite{lin2017focal}, Mask R-CNN~\cite{he2017mask}, Cascade Mask R-CNN~\cite{cai2018cascade}, ATSS~\cite{zhang2020bridging}, GFL~\cite{li2020generalized}, and Sparse R-CNN~\cite{sun2020sparse}.
	Before training, we use the weights pre-trained on ImageNet to initialize the backbone and Xavier~\cite{glorot2010understanding} to initialize the newly added layers.
	We train all the models with  batch size 16 on 8 V100 GPUs, and adopt AdamW~\cite{loshchilov2017decoupled} with an initial learning rate of $1\times10^{-4}$ as optimizer.
	Following common practices~ \cite{lin2017focal,he2017mask,chen2019mmdetection}, we adopt 1$\times$ or 3$\times$ training schedule~(\ie, 12 or 36 epochs) to train all detection models.
	The training image is resized to have a shorter side of 800 pixels, while the longer side does not exceed 1,333 pixels.
	When using the 3$\times$ training schedule, we randomly resize the shorter side of the input image within the range of $[640, 800]$.
	In the testing phase, the shorter side of the input image is fixed to 800 pixels.

	\noindent\textbf{Results.}
	As reported in Tab.~\ref{tab:det_base}, PVT v2 significantly outperforms PVT v1 on both one-stage and two-stage object detectors with similar model size.
	For example, PVT v2-B4 archive 46.1 AP on top of RetinaNet~\cite{lin2017focal}, and 47.5 AP$^{\rm b}$ on top of Mask R-CNN~\cite{he2017mask}, surpassing the models with PVT v1 by 3.5 AP and 4.6 AP$^{\rm b}$, respectively.
	We present some qualitative object detection and instance segmentation results on COCO \texttt{val2017}~\cite{lin2014microsoft} in Fig. \ref{fig:res}, which also shows the good performance of our models.

	For a fair comparison between PVT v2 and Swin Transformer~\cite{swin}, we keep all settings the same, including ImageNet-1K pre-training and COCO fine-tuning strategies.
	We evaluate Swin Transformer and PVT v2 on four state-of-the-arts detectors, including Cascade R-CNN~\cite{cai2018cascade}, ATSS~\cite{zhang2020bridging}, GFL~\cite{li2020generalized}, and Sparse R-CNN~\cite{sun2020sparse}. We see PVT v2 obtain much better AP than Swin Transformer among all the detectors, showing its better feature representation ability.
	For example, on ATSS, PVT v2 has similar parameters and flops compared to Swin-T, but PVT v2 achieves 49.9 AP, which is 2.7 higher than Swin-T. Our PVT v2-Li can largely reduce the computation from 258 to 194 GFLOPs, while only sacrificing a little performance.
	
	\begin{table*}[t]
		\centering
		\setlength{\tabcolsep}{1.7mm}
		\footnotesize

\begin{tabular}{l|c|lcc|lcc|c|lcc|lcc}
\renewcommand{\arraystretch}{0.1}
\multirow{2}{*}{Backbone} &\multicolumn{7}{c|}{RetinaNet 1$\times$} &\multicolumn{7}{c}{Mask R-CNN 1$\times$} \\
\cline{2-15} 
& \#P (M) &AP &AP$_{50}$ &AP$_{75}$ &AP$_S$ &AP$_M$ &AP$_L$ &  \#P (M) &AP$^{\rm b}$ &AP$_{50}^{\rm b}$ &AP$_{75}^{\rm b}$  &AP$^{\rm m}$ &AP$_{50}^{\rm m}$ &AP$_{75}^{\rm m}$\\
\whline
PVTv2-B0 & \textbf{13.0}& \textbf{37.2} & \textbf{57.2} & \textbf{39.5} & \textbf{23.1} & \textbf{40.4} & \textbf{49.7} &\textbf{23.5} & \textbf{38.2} & \textbf{60.5} &\textbf{40.7} & \textbf{36.2} & \textbf{57.8} & \textbf{38.6}  \\
\hline
ResNet18~\cite{he2016deep} &\textbf{21.3} & 31.8 & 49.6 & 33.6 & 16.3 & 34.3 & 43.2  &\textbf{31.2} & 34.0 & 54.0 & 36.7 & 31.2 & 51.0 & 32.7\\
PVTv1-Tiny~\cite{pvt} &23.0& {36.7}& {56.9}& {38.9}& {22.6}& {38.8} &{50.0}  &32.9 & {36.7} & {59.2} & {39.3} & {35.1} & {56.7} & {37.3} \\
PVTv2-B1 (ours) &23.8 & \textbf{41.2} & \textbf{61.9} & \textbf{43.9} & \textbf{25.4} & \textbf{44.5} & \textbf{54.3} &33.7 & \textbf{41.8} & \textbf{64.3} & \textbf{45.9} & \textbf{38.8} & \textbf{61.2} & \textbf{41.6} \\
\hline
ResNet50~\cite{he2016deep} &37.7 & 36.3 & 55.3 & 38.6 & 19.3 & 40.0 & 48.8 & 44.2&38.0 & 58.6 & 41.4 & 34.4 & 55.1 & 36.7\\
PVTv1-Small~\cite{pvt} & {34.2} & {40.4} & {61.3} & {43.0} & {25.0} & {42.9} & {55.7} &{44.1} &{40.4} & {62.9} & {43.8} & {37.8} & {60.1} & {40.3}\\
PVTv2-B2-Li (ours) &\textbf{32.3} & 43.6 &64.7 & 46.8 & 28.3 & 47.6 & 57.4 &\textbf{42.2} & 44.1 &66.3 & 48.4 & 40.5 & 63.2 & 43.6 \\
PVTv2-B2 (ours) &35.1 & \textbf{44.6} & \textbf{65.6} & \textbf{47.6} & \textbf{27.4} & \textbf{48.8} & \textbf{58.6} &45.0& \textbf{45.3} & \textbf{67.1} & \textbf{49.6} & \textbf{41.2} & \textbf{64.2} &\textbf{44.4} \\
\hline
ResNet101~\cite{he2016deep} &56.7  & 38.5 & 57.8 & 41.2 & 21.4 & 42.6 & 51.1  &63.2 & 40.4 & 61.1 & 44.2 & 36.4 & 57.7 & 38.8 \\
ResNeXt101-32x4d~\cite{xie2017aggregated} &56.4& 39.9 & 59.6 & 42.7 & 22.3 & 44.2 & 52.5 &\textbf{62.8} & 41.9 & 62.5 & {45.9} & 37.5 & 59.4 & 40.2 \\
PVTv1-Medium~\cite{pvt} &\textbf{53.9} & {41.9} & {63.1} & {44.3} & {25.0} & {44.9} & {57.6} &63.9 & {42.0} &{64.4} &45.6 &{39.0}& {61.6}& {42.1}\\
PVTv2-B3 (ours) &55.0&\textbf{45.9} & \textbf{66.8} & \textbf{49.3} & \textbf{28.6} & \textbf{49.8} & \textbf{61.4} & 64.9& \textbf{47.0} & \textbf{68.1} & \textbf{51.7} &\textbf{42.5} & \textbf{65.7} &\textbf{45.7} \\
\hline
PVTv1-Large~\cite{pvt} & \textbf{71.1} & {42.6} & {63.7} & {45.4} & {25.8} & {46.0} & {58.4} & \textbf{81.0}& {42.9}& {65.0} & 46.6 &{39.5}& {61.9}& {42.5}  \\
PVTv2-B4 (ours) &72.3&\textbf{46.1} & \textbf{66.9} & \textbf{49.2} & \textbf{28.4} & \textbf{50.0} & \textbf{62.2} &82.2& \textbf{47.5} &\textbf{68.7} & \textbf{52.0} & \textbf{42.7} & \textbf{66.1} &\textbf{46.1} \\
\hline
ResNeXt101-64x4d~\cite{xie2017aggregated} & 95.5& 41.0 & 60.9 & 44.0 & 23.9 & 45.2 & 54.0 &101.9 & 42.8 & 63.8 & {47.3} & 38.4 & 60.6 & 41.3 \\
PVTv2-B5 (ours) &\textbf{91.7} &\textbf{46.2} & \textbf{67.1} & \textbf{49.5} & \textbf{28.5} & \textbf{50.0} & \textbf{62.5} &\textbf{101.6}&\textbf{47.4} &\textbf{68.6} & \textbf{51.9} & \textbf{42.5} & \textbf{65.7}& \textbf{46.0}\\
\end{tabular}
		\caption{\textbf{Object detection and instance segmentation on COCO \texttt{val2017}.} 
			``\#P'' refers to parameter number. AP$^{\rm b}$ and AP$^{\rm m}$ denote bounding box AP and mask AP, respectively. 
			``-Li'' denotes PVT v2 with linear SRA.
		}
		\label{tab:det_base} 
	\end{table*}
	
	\begin{table}[t]
		\centering
		\setlength{\tabcolsep}{0.4mm}
		\footnotesize
		\begin{tabular}{l|c|ccc|ccc}
Backbone &Method & AP$^\text{b}$ & AP$^\text{b}_\text{50}$ & AP$^\text{b}_\text{75}$ & \#P (M) & GFLOPs \\
\whline
 ResNet50~\cite{he2016deep} & \multirow{4}{*}{\makecell{Cascade\\Mask\\R-CNN~\cite{cai2018cascade}}} & 46.3 & 64.3 & 50.5 & 82 & 739  \\
 Swin-T~\cite{swin} & & 50.5 & 69.3 & 54.9 & 86 & 745  \\
PVTv2-B2-Li (ours) & & 50.9 & 69.5 & 55.2 & \textbf{80} &  \textbf{725} \\
PVTv2-B2 (ours) & & \textbf{51.1} & \textbf{69.8} & \textbf{55.3} & 83 &  788 \\
\hline
ResNet50~\cite{he2016deep} & \multirow{4}{*}{ATSS~\cite{zhang2020bridging}} & 43.5	& 61.9 & 47.0 & 32 & 205  \\
Swin-T~\cite{swin} & & 47.2 & 66.5 & 51.3 & 36 & 215  \\
PVTv2-B2-Li (ours) & &  48.9 & 68.1 & 53.4 & \textbf{30}& \textbf{194} \\
PVTv2-B2 (ours) & & \textbf{49.9} & \textbf{69.1} & \textbf{54.1} &33  &  258  \\

\hline
ResNet50~\cite{he2016deep} & \multirow{4}{*}{GFL~\cite{li2020generalized}} & 44.5 & 63.0 & 48.3 &32 & 208  \\
Swin-T~\cite{swin} & & 47.6 &66.8& 51.7 &36 & 215 \\	
PVTv2-B2-Li (ours) & &49.2 & 68.2 & 53.7 & \textbf{30} & \textbf{197}\\
PVTv2-B2 (ours) & & \textbf{50.2} & \textbf{69.4} & \textbf{54.7}  & 33& 261   \\

 \hline
ResNet50~\cite{he2016deep}& \multirow{4}{*}{\makecell{Sparse\\R-CNN~\cite{sun2020sparse}}}& 44.5 & 63.4 & 48.2 & 106 & 166  \\
Swin-T~\cite{swin} & & 47.9 & 67.3 & 52.3 & 110 & 172  \\
PVTv2-B2-Li (ours) & & 48.9 & 68.3 & 53.4 & \textbf{104} & \textbf{151} \\
PVTv2-B2 (ours) & & \textbf{50.1} & \textbf{69.5} &  \textbf{54.9} &  107 & 215  \\
\end{tabular}
		\caption{\textbf{Compare with Swin Transformer on object detection.}
			``AP$^{\rm b}$'' denotes bounding box AP. 
			``\#P'' refers to parameter number.
			``GFLOPs'' is calculated under the input scale of $1280\times 800$. ``-Li'' denotes PVT v2 with linear SRA.
		}
		\label{tab:det_bas2} 
	\end{table}
	
	\subsection{Semantic Segmentation}\label{sec:seg}
	
	\begin{table}[t]
		\centering
		\setlength{\tabcolsep}{1.9mm}
		\footnotesize
		\begin{tabular}{l|c|c|c}
\renewcommand{\arraystretch}{0.1}
	\multirow{2}{*}{Backbone} & \multicolumn{3}{c}{Semantic FPN}\\
	\cline{2-4}
	& \#Param (M) & GFLOPs & mIoU (\%)   \\
	\whline
	PVTv2-B0 (ours) & \textbf{7.6} & \textbf{25.0} & \textbf{37.2} \\
	\hline
	ResNet18~\cite{he2016deep} & \textbf{15.5} & \textbf{32.2} & 32.9 \\
	PVTv1-Tiny~\cite{pvt} & 17.0 & 33.2 & {35.7} \\
	PVTv2-B1 (ours) & 17.8 & 34.2 & \textbf{42.5} \\
	\hline
ResNet50~\cite{he2016deep} & 28.5&45.6 & 36.7\\
PVTv1-Small~\cite{pvt} & {28.2}& 44.5& {39.8}\\
PVTv2-B2-Li (ours) & \textbf{26.3} & \textbf{41.0} & 45.1 \\
PVTv2-B2 (ours) & 29.1 & 45.8 & \textbf{45.2} \\
\hline
ResNet101~\cite{he2016deep} & 47.5&65.1& 38.8\\
ResNeXt101-32x4d~\cite{xie2017aggregated} & \textbf{47.1} &64.7& 39.7 \\
PVTv1-Medium~\cite{pvt} & 48.0 &\textbf{61.0}& {41.6}\\
PVTv2-B3 (ours) & 49.0 & 62.4 & \textbf{47.3} \\
\hline
PVTv1-Large~\cite{pvt} & \textbf{65.1} &\textbf{79.6}& {42.1} \\
PVTv2-B4 (ours) & 66.3 & 81.3 & \textbf{47.9} \\
\hline
ResNeXt101-64x4d~\cite{xie2017aggregated} & 86.4 &103.9& 40.2\\
PVTv2-B5 (ours) & \textbf{85.7} & \textbf{91.1} & \textbf{48.7} \\
\end{tabular}
		\caption{\textbf{Semantic segmentation performance of different backbones on the ADE20K validation set.}
			``GFLOPs'' is calculated under the input scale of $512\times 512$.
			``-Li'' denotes PVT v2 with linear SRA.
		}
		\label{tab:seg}
	\end{table}
	
	\noindent\textbf{Settings.}
	Following PVT v1~\cite{pvt}, we choose ADE20K~\cite{zhou2017scene} to benchmark the performance of semantic segmentation.
	For a fair comparison, we test the performance of PVT v2 backbones by applying it to Semantic FPN~\cite{kirillov2019panoptic}.
	In the training phase, the backbone is initialized with the weights pre-trained on ImageNet~\cite{deng2009imagenet}, and the newly added layers are initialized with Xavier~\cite{glorot2010understanding}. 
	We optimize our models using AdamW~\cite{loshchilov2017decoupled} with an initial learning rate of 1e-4.
	Following common practices~\cite{kirillov2019panoptic,chen2017deeplab}, we train our models for 40k iterations with a batch size of 16 on 4 V100 GPUs.
	The learning rate is decayed following the polynomial decay schedule with a power of 0.9.
	We randomly resize and crop the image to $512\times 512$ for training, and rescale to have a shorter side of 512 pixels during testing.

	\begin{figure*}[t]
		\centering
		\setlength{\fboxrule}{0pt}
		\fbox{\includegraphics[width=0.95\textwidth]{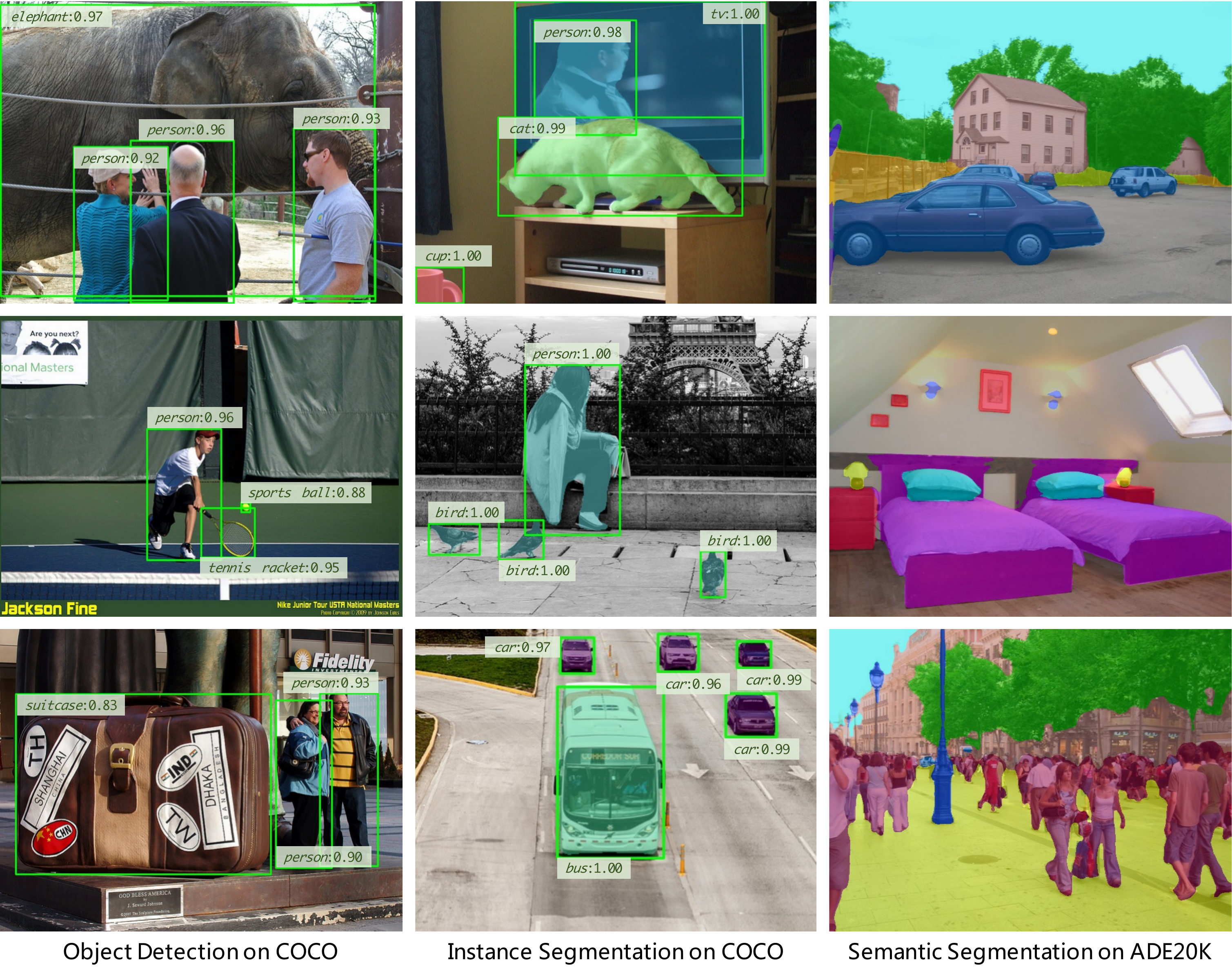}}
		\caption{\textbf{Qualitative results of object detection and instance segmentation on COCO \texttt{val2017}~\cite{lin2014microsoft}, and semantic segmentation on ADE20K~\cite{zhou2017scene}.} The results (from left to right) are generated by PVT v2-B2-based RetinaNet~\cite{lin2017focal}, Mask R-CNN~\cite{he2017mask}, and Semantic FPN~\cite{kirillov2019panoptic}, respectively.}
		\label{fig:res}
	\end{figure*}
	
	\noindent\textbf{Results.} 
	As shown in Tab.~\ref{tab:seg}, when using Semantic FPN~\cite{kirillov2019panoptic} for semantic segmentation, PVT v2 consistently outperforms PVT v1~\cite{pvt} and other counterparts.
	For example, with almost the same number of parameters and GFLOPs, PVT v2-B1/B2/B3/B4 are at least 5.3\% higher than PVT v1-Tiny/Small/Medium/Large. 
	Moreover, although the GFLOPs of PVT-Large are 12\% lower than those of ResNeXt101-64x4d, the mIoU is still 8.5 points higher~(48.7 \textit{vs} 40.2).
	In Fig. \ref{fig:res}, we also visualize some qualitative semantic segmentation results on ADE20K~\cite{zhou2017scene}.
	These results demonstrate that PVT v2 backbones can extract powerful features for semantic segmentation, benefiting from the improved designs.
	
	\subsection{Ablation Study}
	
	\subsubsection{Model Analysis}
	Ablation experiments of PVT v2 is reported in Tab. \ref{tab:model_a}.
	We see that all three designs can improve the model in terms of performance, parameter number, or computation overhead.
	
	\noindent\textbf{Overlapping patch embedding (OPE) is important.} Comparing \#1 and \#2 in Tab. \ref{tab:model_a}, 
	the model with OPE obtains better top-1 accuracy (81.1\% \vs 79.8\%) on ImageNet and better AP (42.2\% \vs 40.4\%) on COCO than the one with original patch embedding (PE)~\cite{dosovitskiy2020image}. OPE is effective because it can model the local continuity of images and feature map via the overlapping sliding window.
	
	\noindent\textbf{Convolutional feed-forward network (CFFN) matters.} Compared to original feed-forward network (FFN)~\cite{dosovitskiy2020image}, our CFFN contains a zero-padding convolutional layer. which can capture the local continuity of the input tensor. 
	In addition, due to the positional information introduced by zero-padding in OPE and CFFN, we can remove the fixed-size positional embeddings used in PVT v1,
	making the model flexible to handle variable resolution inputs.
	As reported in \#2 and \#3 in Tab. \ref{tab:model_a}, CFFN brings 0.9 points improvement on ImageNet (82.0\% \vs  81.1\%) and 2.4 points improvement on COCO, which demonstrates its effectiveness.
	
	\noindent\textbf{Linear SRA (LSRA) contributes to a better model.}
	As reported in \#3 and \#4 in Tab. \ref{tab:model_a}, compared to SRA~\cite{pvt}, our LSRA significantly reduces the computation overhead (GFLOPs) of the model by 22\%, while keeping a comparable top-1 accuracy on ImageNet (82.1\% \vs 82.0\%), and only 1 point lower AP on COCO (43.6 \vs 44.6).
	These results show the low computational cost and good effect of LSRA.
	
	\begin{table}[t]
		\centering
		\setlength{\tabcolsep}{0.8mm}
		\footnotesize
		\begin{tabular}{c|l|c|ccc}
    \renewcommand{\arraystretch}{0.1}
	\multirow{2}{*}{\#} & \multirow{2}{*}{Setting} & \multirow{2}{*}{\tabincell{c}{Top-1 \\Acc (\%)}} & \multicolumn{3}{c}{RetinaNet 1x}\\
	\cline{4-6}
	& & & \#P (M) & GFLOPs &AP  \\
	\whline
	1 & PVTv1-Small~\cite{pvt} & 79.8 & 34.2 & 285.8 & 40.4 \\
	\hline
	2&+ OPE & 81.1 & 34.9 & 288.6 & 42.2 \\
	3&++ CFFN (PVTv2-B2) & 82.0 & 35.1 & 290.7 & \textbf{44.6} \\
	4&+++ LSRA (PVTv2-B2-Li) & \textbf{82.1} & \textbf{32.3} & \textbf{227.4} & 43.6\\
\end{tabular}
		\caption{\textbf{Ablation experiments of PVT v2.} ``OPE'', ``CFFN'', and ``LSRA'' represent overlapping patch embedding, convolutional feed-forward network, and linear SRA, respectively.
		}
		\label{tab:model_a}
	\end{table}
	
	\subsubsection{Computation Overhead Analysis}
\begin{figure}
		\centering
		\setlength{\fboxrule}{0pt}
		\fbox{\includegraphics[width=0.45\textwidth]{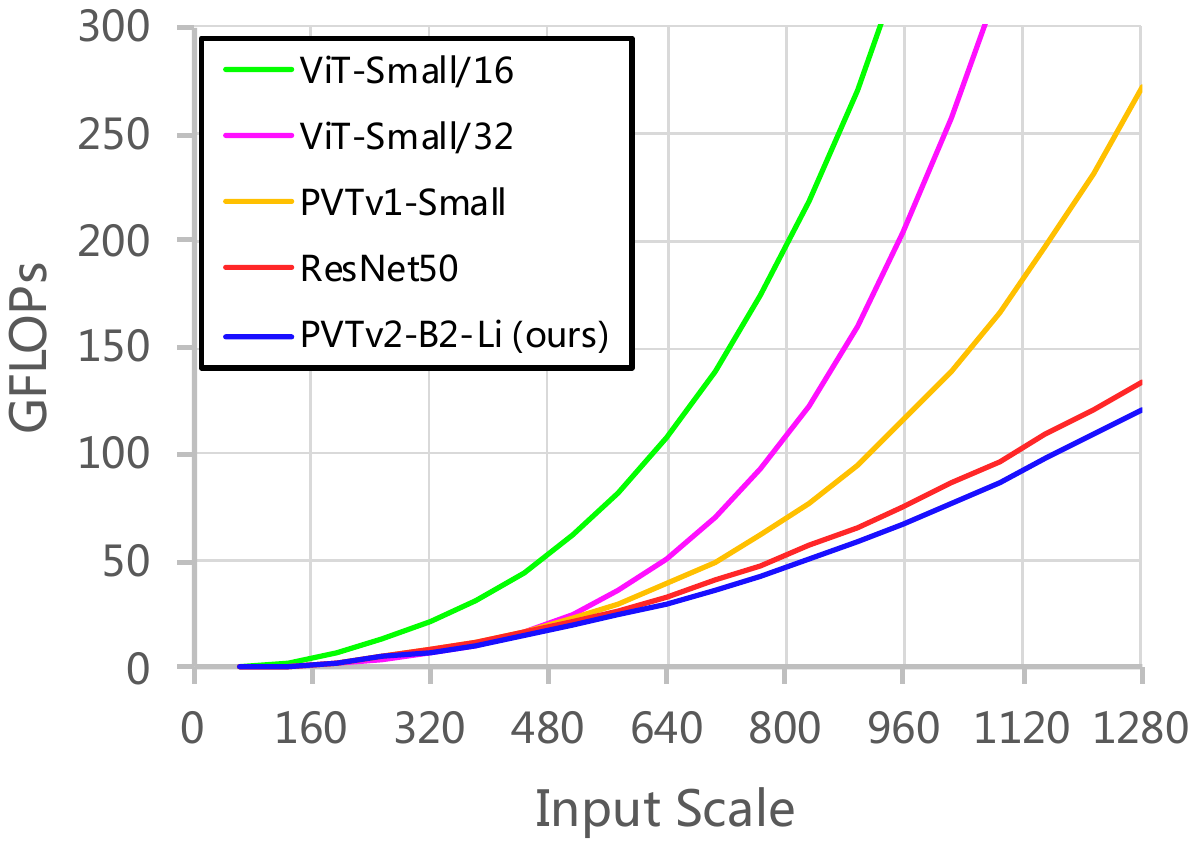}}
		\vspace{-10pt}
		\caption{\textbf{Models' GFLOPs under different input scales.} The growth rate of GFLOPs: ViT-Small/16~\cite{dosovitskiy2020image}$>$ViT-Small/32~\cite{dosovitskiy2020image}$>$PVT v1-Small~\cite{pvt}$>$ResNet50~\cite{he2016deep}$>$PVT v2-B2-Li (ours).
		}
		\label{fig:flops}
	\end{figure}

	As shown in Figure \ref{fig:flops}, with increasing input scale, the GFLOPs growth rate of the proposed PVT v2-B2-Li is much lower than that of PVT v1-Small~\cite{pvt}, and is similar to that of ResNet-50~\cite{he2015delving}.
	This result proves that our PVT v2-Li successfully addresses the high computational overhead problem caused by the attention layer.
	
	\section{Conclusion}
	
	We study the limitations of Pyramid Vision Transformer (PVT v1) and improve it with three designs, which are overlapping patch embedding, convolutional feed-forward network, and linear spatial reduction attention layer.
	Extensive experiments on different tasks, such as image classification, object detection, and semantic segmentation demonstrate that the proposed PVT v2 is stronger than its predecessor PVT v1 and other state-of-the-art transformer-based backbones, under comparable numbers of parameters.
	We hope these improved baselines will provide a reference for future research in vision Transformer.
	
	{\small
		\bibliographystyle{ieee_fullname}
		\bibliography{egbib}
	}
	
\end{document}